\documentclass[sigconf]{acmart}

\AtBeginDocument{%
  \providecommand\BibTeX{{%
    \normalfont B\kern-0.5em{\scshape i\kern-0.25em b}\kern-0.8em\TeX}}}

\usepackage{booktabs} 

\setcopyright{rightsretained}
\copyrightyear{2019}
\acmYear{2019}
\acmDOI{}

\acmConference[VRST '19]{25th ACM Symposium on Virtual Reality Software and Technology}{November 12--15, 2019}{Parramatta, NSW, Australia}
\acmBooktitle{25th ACM Symposium on Virtual Reality Software and Technology (VRST '19), November 12--15, 2019, Parramatta, NSW, Australia}
\acmPrice{15.00}
\acmISBN{978-1-4503-7001-1/19/11}
 
\begin{document}
\title{Development of MirrorShape: High Fidelity Large-Scale Shape Rendering Framework for Virtual Reality}
\author{Aleksey Fedoseev}
\affiliation{%
  \institution{Skolkovo Institute of Science and Technology (Skoltech)}
  \streetaddress{Nobelya Ulitsa 3}
  \city{Moscow}
  \country{Russia}
  \postcode{121205}
}
\email{aleksey.fedoseev@skoltech.ru}

\author{Nikita Chernyadev}
\affiliation{%
  \institution{Skolkovo Institute of Science and Technology (Skoltech)}
  \streetaddress{Nobelya Ulitsa 3}
  \city{Moscow}
  \country{Russia}
  \postcode{121205}
}
\email{nikita.chernyadev@skoltech.ru}

\author{Dzmitry Tsetserukou}
\affiliation{%
  \institution{Skolkovo Institute of Science and Technology (Skoltech)}
  \streetaddress{Nobelya Ulitsa 3}
  \city{Moscow}
  \country{Russia}
  \postcode{121205}
}
\email{d.tsetserukou@skoltech.ru}

\renewcommand{\shortauthors}{Fedoseev, Chernyadev and Tsetserukou}

\begin{abstract}
Today there is a high variety of haptic devices capable of providing tactile feedback. Although most of existing designs are aimed at realistic simulation of the surface properties, their capabilities are limited in attempts of displaying shape and position of virtual objects. 
  
This paper suggests a new concept of distributed haptic display for realistic interaction with virtual object of complex shape by a collaborative robot with shape display end-effector. MirrorShape renders the 3D object in virtual reality (VR) system by contacting the user hands with the robot end-effector at the calculated point in real-time. Our proposed system makes it possible to synchronously merge the position of contact point in VR and end-effector in real world. This feature provides presentation of different shapes, and at the same time expands the working area comparing to desktop solutions.

The preliminary user study revealed that MirrorShape was effective at reducing positional error in VR interactions. Potentially this approach can be used in the virtual systems for rendering versatile VR objects with wide range of sizes with high fidelity large-scale shape experience.

\end{abstract}

%
%

\ccsdesc[500]{Computing methodologies~Virtual reality}
\ccsdesc[500]{Haptic devices}
\ccsdesc[300]{Computer systems organization~Robotics}

\keywords{3D interaction, collaborative technologies, haptics, interaction technologies, robotics, shape-changing interfaces, virtual reality}

\begin{teaserfigure}
  \includegraphics[width=\textwidth]{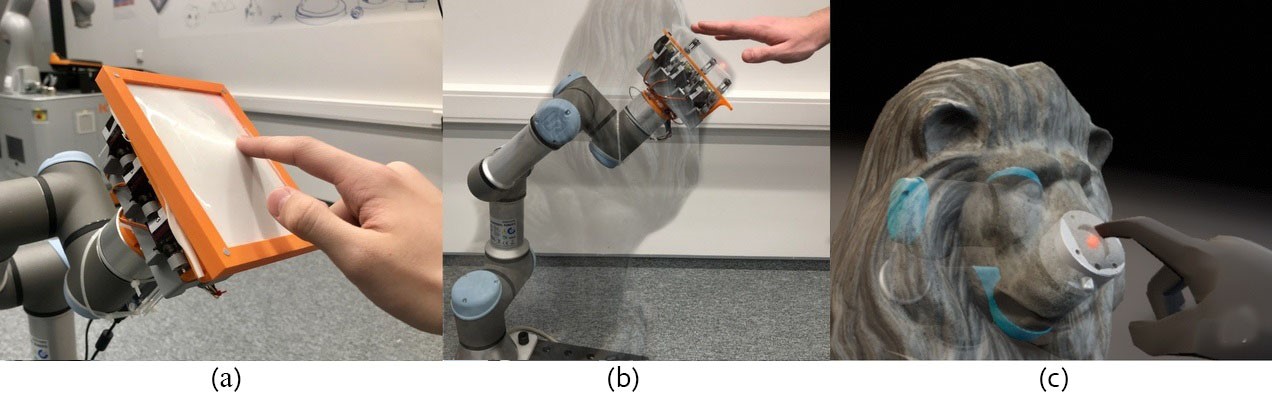}
  \caption{ (a) Human-robot interaction, (b) User approaching the virtual object, (c) User scene with real-time simulation.}
  \label{fig:ill5}
\end{teaserfigure}

\maketitle
\section{Introduction}
Over the past decade, there was a significant development in the scientific field of haptic devices. Still, much attention in modern research is focused on the tactile simulation in tasks of manipulating with small objects. In order to improve the haptic feedback, the developed devices usually take the form of wearable actuated gloves \cite{yem2017wearable}, exoskeletons \cite{tsetserukou2010exointerfaces} or handheld controllers, which allows them to reach greater mobility, but limits the actual positional feedback. Some works in this area, such as NormalTouch and TextureTouch, developed by Benko's research group \cite{Benko:2016:NTH:2984511.2984526} and Mediate developed by Fitzgerald's research group \cite{Fitzgerald:2018:MST:3170427.3188472} are aiming to use a pin matrix that is driven by linear actuators to create the shape of the surface. However, these solutions with high mobility allow to simulate  the normal contact force or the shape of an object without combining these parameters. Additionally, they can generate sensation only in the narrow area, which is not the case in large-scaled VR systems.

Separate attempts to render large objects of complex shape were made, e.g., Snake Charmer developed by the research group of the Toronto University  \cite{Araujo:2016:SCP:2839462.2839484}. Their article presents a novel idea of using manipulator to achieve haptic feedback with an efficient algorithm for rendering large objects of simple shapes. The resolution of this display still remains too low to simulate a change in the surface shape. To transmit both shape and a positional feedback in real time the MirrorShape design aims to use robotic arm along with shape feedback in the contact surface.

\section{Principle and Technologies}
The MirrorShape system software consists of four modules (Figure 2), which includes Hand tracking module, VR environment, Robot controller and end-effector haptic display MirrorGlide. 

\begin{figure} [h]
  \centering
  \includegraphics[width=0.8\linewidth]{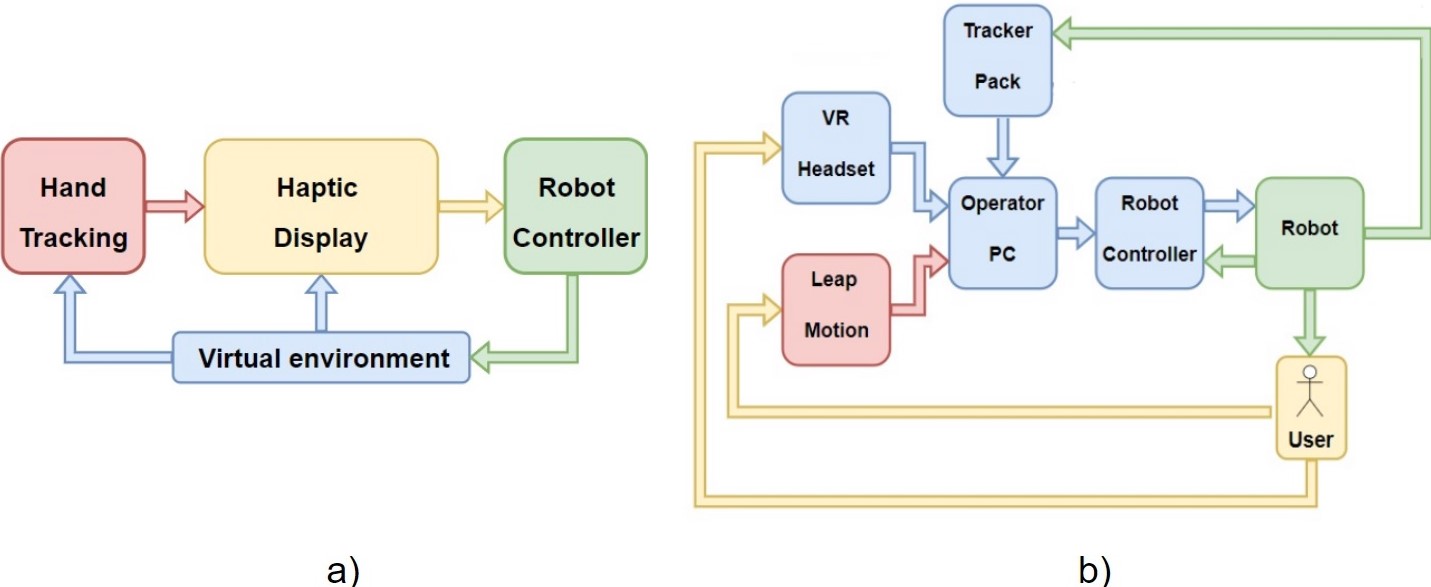}
  \caption{The MirrorShape (a) software and (b) hardware \\
  architecture}
  \label{fig4}
\end{figure}
The system hardware consists of a 6 DOF Universal robot UR3, PC with control framework, tracking system, and shape-forming end-effector (Figure 3) as haptic display. User wears HTC Vive Pro headset with Leap Motion camera to detect the position of the hands. The design of inverted five bar linkage mechanism allows MirrorGlide to define the position of 3 contact points independently, achieving a fairly wide range of the working area. The generation of 3 contact points combined with silicone display provides interaction with various shapes only with 3 DOF device instead of using large number of pins and actuators.

In order to improve the experience of human-robot interaction, a full digital robot twin with 6 DOF was developed and rendered within the Unity Engine as well as the remote control framework for UR robot, which allows the adaptive surface simulation in real-time. The hand position is tracked by Leap Motion Module to determine the coordinates of the user's fingertips. Smoothing algorithm was implemented for tracking data, which allowed to make a calibration of the real robot position in the virtual space environment.

\begin{figure} [h]
  \centering
  \includegraphics[width=0.6\linewidth]{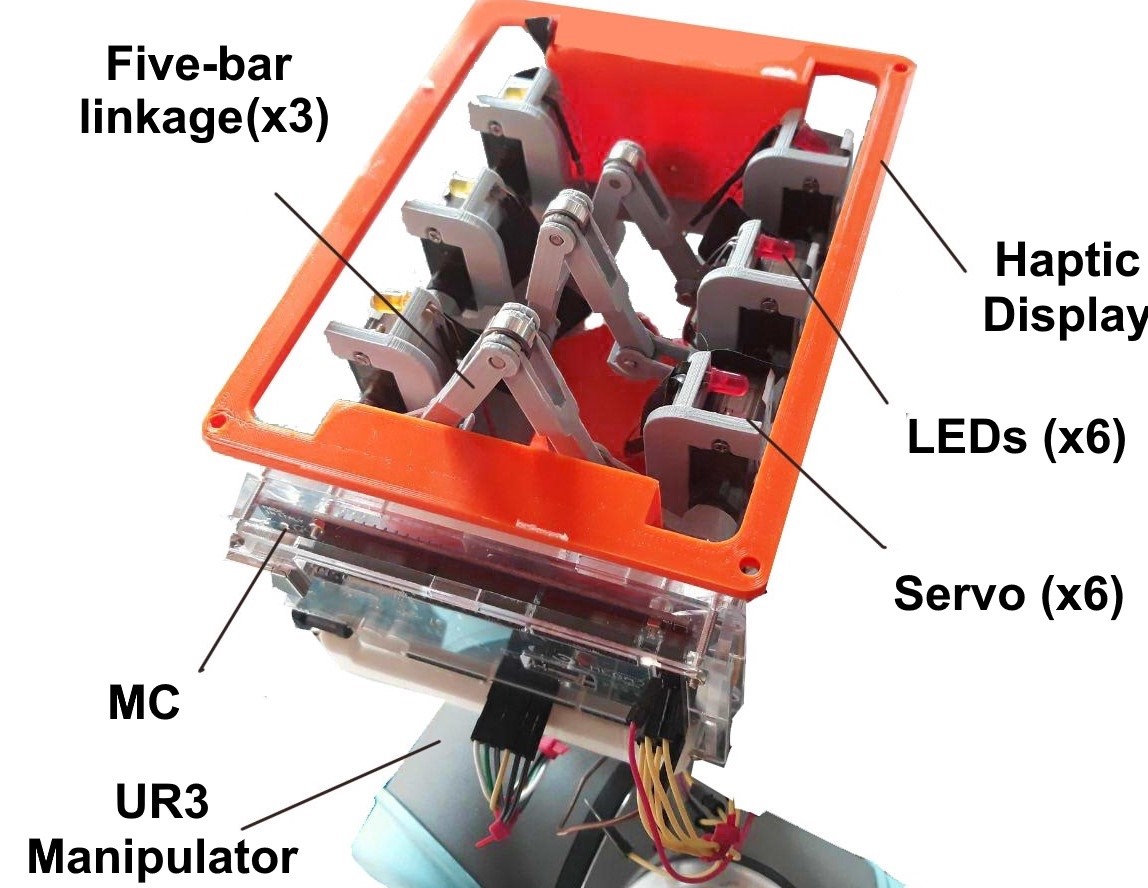}
  \caption{The MirrorGlide display architecture.}
  \label{fig2}
\end{figure}


When user approaches simulated objects, a field of approximate contact with the surface is determined in VR environment, after which the real robot as well as digital twin begins its movement to a predetermined position, in order to be at the same point as the normal vector to the fingertip touching the surface. Depending on the type of the investigated surface and the curvature of the object shape between the hand and the display plane, the end-effector performs forming of various shapes.

\section{Applications and Conclusions}

When rendering interactions with VR surfaces, a large role is played by positional desynchronization of visual and tactile feedback. Experiments on precision were conducted in this study (Figure 4) where 6 participants were asked to measure the platforms of different length. 
\begin{figure} [h]
  \centering
  \includegraphics[width=0.7\linewidth]{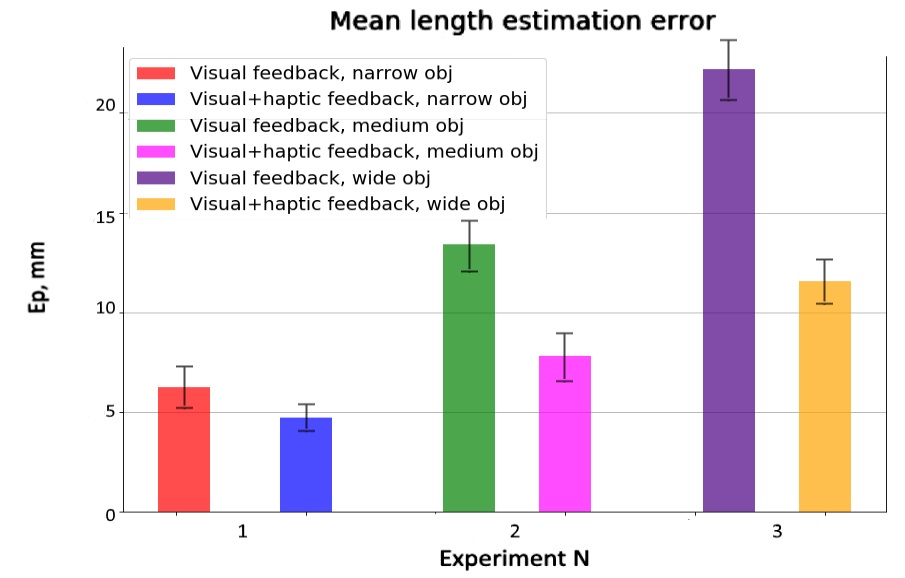}
  \caption{ The experimental results. }
  \label{fig3}
\end{figure}

The each participant had to track two events: the beginning and the end of the interaction with the surface, while moving their hand along the objects. The results of the estimated size errors showed that the MirrorShape allows to decrease positional error in large-scale interactions and potentially can be used for rendering versatile VR objects.  

\bibliographystyle{ACM-Reference-Format}
\bibliography{sample-bibliography}

\end{document}